\documentclass[letterpaper]{article} % DO NOT CHANGE THIS
\usepackage[preprint]{aaai2027}  % arXiv preprint: show authors, omit AAAI copyright
\usepackage[hyphens]{url}  % DO NOT CHANGE THIS
\usepackage{graphicx} % DO NOT CHANGE THIS
\usepackage{natbib}  % DO NOT CHANGE THIS AND DO NOT ADD ANY OPTIONS TO IT
\usepackage{caption} % DO NOT CHANGE THIS AND DO NOT ADD ANY OPTIONS TO IT
\usepackage{algorithm}
\usepackage{algorithmic}
\usepackage{microtype}

\usepackage{newfloat}
\usepackage{listings}
\DeclareCaptionStyle{ruled}{labelfont=normalfont,labelsep=colon,strut=off} % DO NOT CHANGE THIS
\floatstyle{ruled}
\newfloat{listing}{tb}{lst}{}
\floatname{listing}{Listing}
\usepackage{amsmath}

\usepackage{multirow}
\usepackage{array}
\usepackage{CJKutf8}
\usepackage{tabularx}

\makeatletter
\expandafter\let\csname ver@CJK.sty\endcsname\relax
\makeatother

\usepackage{booktabs}

\title{AgenticASR: Refining Speech Recognition in Real-World Scenarios via an Agentic Approach}
\author {
    Zixuan Jiang\textsuperscript{\rm 1,\rm 2,\rm 3}\equalcontrib,
    Binghao Qiang\textsuperscript{\rm 1}\equalcontrib,
    Jiaying Chi\textsuperscript{\rm 1}\equalcontrib,
    Yanqiao Zhu\textsuperscript{\rm 1,\rm 2},
    Kai Yu\textsuperscript{\rm 1},
    Xie Chen\textsuperscript{\rm 1,\rm 2}\corresponding
}
\affiliations {
    \textsuperscript{\rm 1}X-LANCE lab, Shanghai Jiao Tong University\\
    \textsuperscript{\rm 2}Shanghai Innovation Institute\\
    \textsuperscript{\rm 3}College of Artificial Intelligence, Xi'an Jiaotong University\\
    andrewjiang@stu.xjtu.edu.cn, casper.qiang@sjtu.edu.cn, chijiaying@sjtu.edu.cn,
    chenxie95@sjtu.edu.cn
}

\begin{document}

\maketitle

\begin{abstract}
Automatic speech recognition (ASR) has achieved substantial gains in transcription
accuracy, yet verbatim transcription does not necessarily produce readily usable
text. It retains fillers, repetitions, false starts, and
self-corrections that increase reading effort, obscure the speaker's final
intent, and propagate unresolved or abandoned content to downstream tasks. Existing spoken-to-written methods process completed audio or transcripts
but cannot revise emitted text when later speech changes how preceding content
should be interpreted. We therefore formulate Agentic Speech Recognition
(AgenticSR), an audio-to-clean-text task that removes disfluencies, resolves
self-corrections, and normalizes written form while preserving the speaker's
final intent. AgenticASR implements this task through an ASR--Refiner
architecture that repeatedly transforms a bounded active context and replaces
its corresponding output span as audio arrives. This enables continual emission
and revision over streams of arbitrary duration. We also introduce AASR-Bench, a
bilingual benchmark with fine-grained atomic rubrics. Across multiple ASR front
ends, AgenticASR attains the highest AASR-Bench scores among evaluated systems. A
human--AI agreement study shows that rubric-based judgments align with independent
expert assessments. Ablations characterize Refiner capacity, context length, and
the quality--latency trade-off between online and offline inference.
Together, these results establish AgenticASR as a practical framework for
intent-preserving clean transcription during ongoing speech. Code,
AASR-Bench, and a demo will be released at \url{https://github.com/AnXMuy/AgenticASR}.
\end{abstract}


\section{Introduction}

Automatic speech recognition (ASR) underpins voice interfaces by converting
speech into verbatim text for downstream processing \cite{prabhavalkar2024survey}.
Architectural and data-scaling advances have produced strong multilingual
recognition systems \cite{graves2006connectionist,graves2012sequence,
radford2023robust,shi2026qwen3}.

The spoken form captured by a verbatim transcript often differs from the written
form needed for reading and downstream use \cite{biber2019}. Spontaneous speech
contains fillers, repetitions, false starts, self-corrections, and spoken-form
expressions \cite{bortfeld2001disfluency,tan2023four}. Converting such speech
into readable text requires post-processing, including disfluency filtering,
repetition removal, self-correction resolution, inverse text normalization
(ITN), and written-form formatting. Verbatim retention supports lexical-fidelity
evaluation \cite{chen2021gigaspeech}, but can reduce readability and obscure
final intent
\cite{liao2023improving,wang2010disfluency,honnibal2014joint}. In interactive
agents, such artifacts can also propagate into intent detection and dialogue
decisions \cite{dao2022disfluency,marie2023disfluency}.

\begin{figure*}[htbp!]
    \centering
    \includegraphics[width=\textwidth]{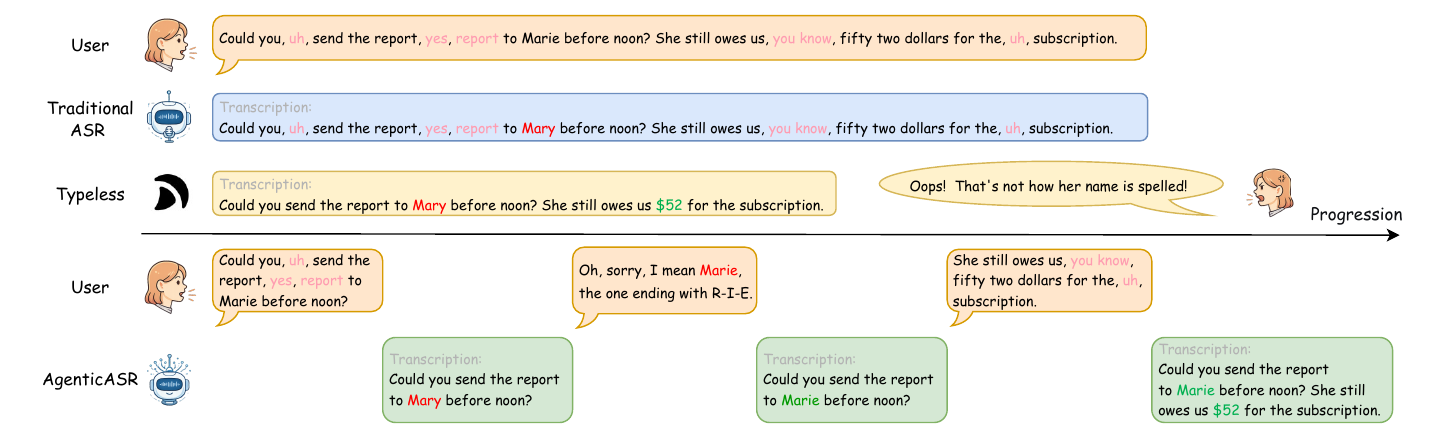}
    \caption{Comparison of verbatim ASR, offline spoken-to-written
    transformation, and online AgenticASR. Traditional ASR preserves the spoken
    surface form, including disfluencies and repetitions. Offline
    post-transformation systems such as Typeless wait until an utterance is
    complete before producing a clean transcript. Consequently, later spoken
    disambiguation, such as specifying that a name ends with ``R-I-E'', cannot
    revise an earlier result within the same ongoing interaction. AgenticASR
    instead emits clean partial text online and revises previously emitted
    content when later speech provides new evidence, as illustrated by the
    update from ``Mary'' to ``Marie''.}
    \label{fig:teaser}
\end{figure*}

As shown in figure \ref{fig:teaser}, existing spoken-to-written systems use cascaded post-transformation,
end-to-end spoken-to-formal ASR, or interactive revision
\cite{typeless,opentypeless,ning2026formalasr,
jiang2026humanlikeinteractivespeechrecognition}. They generally transform
completed audio or transcripts, delaying clean text until an utterance ends.
A
  natural way to reduce this latency is to emit cleaned partial hypotheses as
  speech arrives. However, early emission alone is insufficient because a later
  correction, spelling, or explanation may invalidate text that has already been
  committed. An online system must therefore revise recent output as new
  evidence arrives, without repeatedly processing the entire stream.

We therefore formalize \emph{Agentic Speech Recognition} (AgenticSR) as an
audio-to-clean-text task. Given audio, the desired output is a readable written
transcript of the speaker's final intended message. It removes semantically empty
fillers, repetitions, and abandoned corrections; normalizes spoken expressions
and formatting; preserves all intended content; and leaves already clean input
unchanged. We instantiate AgenticSR with \textbf{AgenticASR}, a two-stage system in
which an ASR front end produces an intermediate hypothesis and a Refiner maps
that hypothesis to clean text. At each update, AgenticASR concatenates the active
recent source text, obtains one refined string, and replaces the corresponding
local span in the emitted transcript. We introduce \textbf{AASR-Bench}, to our
knowledge the first dedicated bilingual benchmark for this task, together with
an LLM-assisted pipeline that constructs training pairs for compact Refiners.

Our contributions are as follows:
\begin{itemize}
    \item We define Agentic Speech Recognition (AgenticSR) as a
    final-intent-preserving audio-to-clean-text recognition task. Valid outputs
    retain the semantic content of the final intended message, make only
    evidence-supported edits, and pass clean input through unchanged.
    \item We introduce \textbf{AASR-Bench}, to our knowledge the first dedicated
    bilingual benchmark for AgenticSR. Its pass-through controls and 6,637
    atomic rubrics separately evaluate Content, Format, Filter, and Rephrase.
    \item We develop \textbf{AgenticASR}, a decoupled ASR-Refiner system for
    online AgenticSR. An LLM-assisted pipeline constructs task-specific training
    pairs, and the Refiner transforms a concatenated, bounded recent source
    context into one clean string that replaces the corresponding local output
    span as speech arrives.
\end{itemize}

\section{Related Work}
\subsection{Automatic Speech Recognition}
Modern ASR systems have advanced through architectural innovation, data scaling,
and integration with large language models. Whisper uses large-scale
multilingual training with a Transformer encoder--decoder, while Qwen3-ASR
combines speech representations with a language-model decoder
\cite{radford2023robust,shi2026qwen3}. These systems are optimized primarily for
verbatim transcription and lexical-fidelity metrics such as WER.

Spoken-to-written conversion treats oral phenomena as material that may be
inappropriate in a written target \cite{ihori2020,guo2023cs2w,liu2025recording}.
Typeless and OpenTypeless apply a post-transformation stage to a recognized
transcript \cite{typeless,opentypeless}. Other approaches restore punctuation,
normalization, or disfluency removal \cite{alam2020punctuation,fu2021improving,
zhang2022capitalization,tan2023four}, and neural correction models address ASR
errors efficiently \cite{leng2021fastcorrect,leng2023softcorrect}. FormalASR
instead fine-tunes Qwen3-ASR to generate formal text directly from speech
\cite{ning2026formalasr}. Interactive ASR can revise transcripts with dialogue
context and user feedback \cite{jiang2026humanlikeinteractivespeechrecognition}.
In contrast, AgenticASR keeps the external AgenticSR task audio-to-clean-text,
while using a backbone-independent Refiner to transform a concatenated active
window of intermediate ASR text into a single replacement string. This separation
supports bounded local revision during ongoing speech without treating chunks as
independent Refiner outputs.

\subsection{Benchmarks for Speech Tasks}
Traditional ASR evaluation primarily relies on lexical matching metrics such as
WER, CER, and MER, which measure transcription accuracy at the token level.

To better capture semantic quality, prior work has proposed semantics-aware
metrics. LLMs have shown strong alignment with human judgments in semantic
evaluation \cite{liu2025recording,liu2023g,zheng2023judging}, motivating their
adoption for ASR evaluation. S2ER uses an LLM to assess sentence-level semantic
preservation \cite{wang2026interactiveasrhumanlikeinteraction}, while AER
measures information preservation by comparing LLM-derived answers from the
reference and transcript \cite{pulikodan2025approach}. MMAE further employs
LLM-based rubrics for fine-grained multidimensional evaluation \cite{ma2026mmae}.

Inspired by these approaches, we introduce \textbf{AASR-Bench}, a dedicated
bilingual benchmark for AgenticSR with rubric-based LLM evaluation. Each rubric
question evaluates an atomic requirement of readable transcription that preserves
final intent, including disfluency removal, self-correction resolution, ITN,
and formatting. This design provides fine-grained, interpretable evaluation for
the AgenticSR task.

\section{Method}

\subsection{AASR-Bench}

\subsubsection{Task Formulation}

Agentic Speech Recognition (AgenticSR) converts an audio input into a clean
written transcript of the speaker's final intended message. Conventional ASR
targets a verbatim transcript and therefore retains each spoken token. In
contrast, AgenticSR resolves spoken-language artifacts to produce the text that a
reader would expect in writing. A valid AgenticSR output should:
\begin{itemize}
    \item remove semantically empty fillers and smooth repetitions or stuttering;
    \item apply ITN to expressions such as numbers and dates, and render entities
    in conventional written forms;
    \item resolve false starts and self-corrections, retaining only the speaker's
    final intended content;
    \item use contextual spelling or explanatory cues to recover uncommon
    entities, for example mapping ``Mary, ends with ie'' to ``Marie.''
\end{itemize}
The output must preserve all content that remains part of the final intended
message. For an utterance that is already suitable for writing, AgenticSR should
return the input unchanged. AgenticASR realizes this task with an ASR front end and a Refiner that
converts the front end's spoken-form hypothesis into clean written text.

\subsubsection{Benchmark Construction}

We construct AASR-Bench as a bilingual benchmark of controlled
\emph{Oral--Clean} pairs across 10 usage scenes: academic, customer service,
daily chat, dictation memo, explanation, meeting, navigation, tech, vibe coding,
and voice search. Together, they cover three axes: interaction mode
(conversation, dictation, or command), lexical domain (daily, professional, or
technical), and transformation demand (disfluency removal, written-form
normalization, or numerical and entity correction). The dedicated
\emph{explanation} scene isolates late spelling and entity-clarification cues.
From online media, we collect naturally occurring Clean sentences that reflect
everyday communication, present moderate difficulty, and provide sufficient
context. We apply ITN and written-form formatting, then remove near duplicates
using text 3-gram Jaccard similarity at a 0.75 threshold.

After deduplication, we manually construct the corresponding Oral sentences.
Depending on the source, we preserve existing spoken phenomena or add fillers,
repetitions, stuttering, a controlled number of self-corrections, explanations,
and spelling cues. Explanations are restricted to the dedicated
\emph{explanation} scene. We cross-check all pairs to verify that each Oral
sentence is valid, each Clean sentence preserves the intended meaning, and the
correspondence is unambiguous.

Each scene includes pass-through cases to detect over-editing that introduces
unsupported or hallucinated content. Most Oral sentences are synthesized with
Doubao TTS 2 \cite{doubao}; the remaining 16.36\% are recorded manually because
complex explanations or self-corrections cannot be reliably expressed by TTS.

\subsubsection{Rubric-Based Evaluation}

Token-level metrics such as WER cannot distinguish valid formatting alternatives
from semantic errors. They also cannot localize failures in preservation,
normalization, or Oral-to-Written transformation. We therefore evaluate each
output using atomic multiple-choice rubrics generated by Qwen3.7-Plus. The
rubrics cover four dimensions:
\textbf{Content} checks whether semantic units unaffected by spoken-language
phenomena are preserved; \textbf{Format} checks ITN and written formatting of
numbers, dates, symbols, and entities without requiring a single surface form;
\textbf{Filter} checks the removal of semantically empty fillers, repetitions,
and parenthetical speech; and \textbf{Rephrase} jointly evaluates
self-corrections, multi-stage revisions, and explanations. For corrections and
revisions, it checks whether the final intended result is retained while
abandoned content and the revision process are removed. For explanation samples,
it checks whether the explained entity is recovered correctly while redundant
explanatory or spelling cues are omitted. Gemma-4-31B-IT
\cite{gemmateam2026gemma4} serves as the judge and selects one option for each
rubric question. All generated rubric questions and answer options were manually
checked for correctness and relevance.

The option scores are $\{1,0,-1\}$ for Content and Format, $\{2,0\}$ for
Filter, and $\{2,1,0,-1\}$ for Rephrase. Following the positive--negative
question design of DLC-Bench \cite{lian2025describe}, negative scores penalize
missing, contradictory, or hallucinated information and increase the separation
between faithful transformation and harmful editing. Let $s_{ij}$ be the score of
question $j$ for sample $i$, and let $s^{\max}_{ij}$ be its maximum possible
score. We report the question-weighted micro-average in percentage points:
\begin{equation}
    S = 100 \cdot \mathrm{clip}_{[0,1]}
    \left(
    \frac{\sum_i \sum_j s_{ij}}
         {\sum_i \sum_j s^{\max}_{ij}}
    \right),
\end{equation}
Category and scene-level scores use the same aggregation. All systems are
evaluated on the full benchmark. For an ASR request that remains unsuccessful
after three attempts, we retain an empty output and score it with the same
rubrics rather than excluding it. All judge requests completed successfully in
the reported runs; no judge result is excluded from the aggregates.

% \begin{figure*}[htbp!]
%     \centering
%     \includegraphics[width=\textwidth]{Figures/AgenticASR-method1.pdf}
%     \caption{Overview of AgenticASR. (a) The data-generation pipeline constructs spoken-form and clean-text training pairs. (b) Offline inference
%   refines a complete ASR hypothesis after the utterance ends. For online inference, the Chunk Manager selects $C_t$ and up to $K-1$ preceding spans,
%   concatenates their ASR text as $W_t$, and sends it to the Refiner. The resulting clean-text string replaces the previous output for the selected
%   window; as the window advances, later context can revise earlier output. We use $K=3$ by default.}
%     \label{fig:method-overview}
% \end{figure*}

\begin{figure*}[htbp!]
    \centering
    \includegraphics[width=\textwidth]{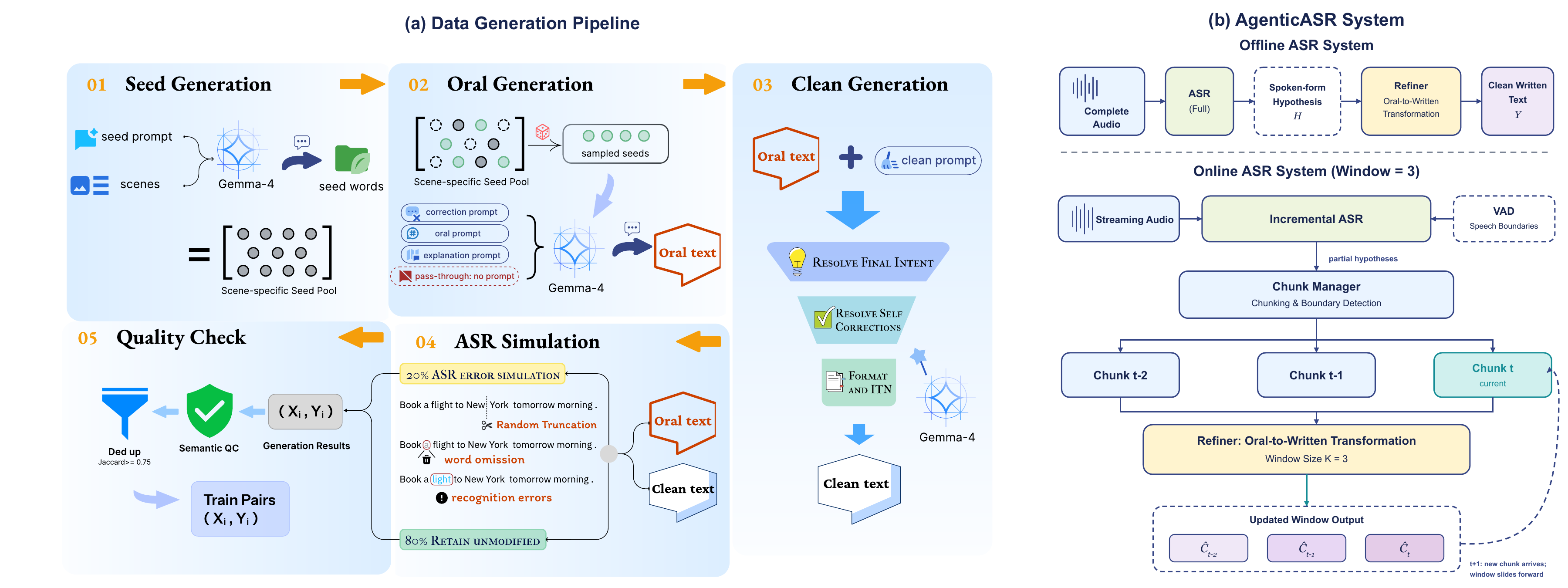}
    \caption{Overview of AgenticASR. (a) The five-stage data pipeline generates
    Refiner training pairs through seed, Oral, and Clean generation, ASR
    simulation, quality control, and deduplication. (b) The AgenticASR workflow
    supports both offline and online inference. Offline inference refines the
    complete ASR hypothesis once. Online inference uses VAD and the Chunk Manager
    to maintain a sliding $K$-chunk window; each refinement replaces its
    clean-text span, allowing later speech to revise earlier output. We use
    $K=3$ by default.}
    \label{fig:method-overview}
\end{figure*}

\subsubsection{Benchmark Statistics}

AASR-Bench contains 917 utterances covering 10 usage scenes: academic,
customer service, daily chat, dictation memo, explanation, meeting, navigation,
tech, vibe coding, and voice search. The 510 Chinese and 407 English samples
total 4.218 hours of audio from 10 voice identities, including two human
speakers. Of the 917 samples, 129 are pass-through controls (14.07\%), and 47
belong to the dedicated explanation scene, the only scene containing explanation
phenomena.

The benchmark comprises 6,637 rubric questions, averaging 7.24 per sample
(range: 2--19). Table~\ref{tab:aasrbench-rubrics} reports their distribution.
Every sample has at least one Content question; other dimensions are included
only when applicable to the phenomena present in that sample.
\begin{table}[htbp!]
    \centering
    \small
    \begin{tabular}{lrrr}
        \toprule
        Category & Questions & Share (\%) & Coverage \\
        \midrule
        Content  & 3,448 & 51.95 & 917 \\
        Format   & 1,498 & 22.57 & 741 \\
        Filter   &   882 & 13.29 & 882 \\
        Rephrase &   809 & 12.19 & 623 \\
        \midrule
        Total    & 6,637 & 100.00 & -- \\
        \bottomrule
    \end{tabular}
    \caption{Distribution of atomic rubrics in AASR-Bench. Coverage denotes the
    number of samples containing at least one rubric of that category.}
    \label{tab:aasrbench-rubrics}
\end{table}

\subsection{AgenticASR}

AgenticASR is a two-stage system comprising an ASR module and an LM-based
Refiner. The Refiner implements the Oral-to-Written transformation over ASR
hypotheses. Online inference repeatedly selects a bounded recent source span,
concatenates its text, and replaces the corresponding local written span with
one refined string. Figure~\ref{fig:method-overview} summarizes the complete
method. We first present the LLM-assisted data-generation pipeline used to train
the Refiner and then describe offline and online inference.

% Figure source note: the online path is Audio Stream -> ASR -> Chunk Manager
% -> concatenated active source context -> Compact LM Refiner -> one refined
% text string -> replacement of the corresponding local written span. Chunks
% identify source boundaries and replacement scope; they are not model slots.
\subsubsection{Data Generation Pipeline}

To train the Refiner for Oral-to-Written transformation, we use an LLM-assisted
data-generation pipeline that constructs ASR-style
input--Clean-target pairs through intermediate Oral--Clean pairs. These pairs
train the Refiner used in AgenticASR; they do not alter the audio-to-clean-text
definition of AgenticSR. All LLM-based generation and quality-control operations
use Gemma-4-31B-IT \cite{gemmateam2026gemma4} with stage-specific prompts. The
pipeline consists of five stages.

We generate training data for the same 10 scenes used in AASR-Bench. For a
target size $N$, we allocate examples across these scenes according to
Table~\ref{tab:training-scene-mixture}. The ratios are applied before filtering
and deduplication.

\begin{table}[htbp!]
    \centering
    \small
    \begin{tabular}{lrlr}
        \toprule
        Scene & \% & Scene & \% \\
        \midrule
        Daily chat       & 15 & Vibe coding    & 12 \\
        Explanation      & 12 & Meeting        & 11 \\
        Customer service & 11 & Academic       &  9 \\
        Navigation       &  3 & Dictation memo & 10 \\
        Voice search     &  7 & Tech           & 10 \\
        \bottomrule
    \end{tabular}
    \caption{Target scene allocation for data generation. Ratios are specified
    before semantic quality control and global deduplication.}
    \label{tab:training-scene-mixture}
\end{table}

\paragraph{Stage 1: Seed generation.}
For each scene, the Seed Prompt instructs the LLM to construct a pool
$\mathcal{C}_s$ of relevant entities, terms, numerical patterns, and long-tail
expressions. Later rounds include existing seeds in an exclusion list to reduce
duplication. Before each pass, we shuffle the seed pool. For each sample $i$ from
its corresponding scene $s_i$, we then take five consecutive seeds from the
shuffled pool and denote this candidate set by $\mathcal{S}_i$.

\paragraph{Stage 2: Oral generation.}
Given scene $s_i$ and five candidate seeds $\mathcal{S}_i$, the LLM selects
three to five seeds and composes a coherent Oral utterance. Two independent
prompts control its realization:
\begin{equation}
    (\widetilde{\mathcal{S}}_i,T_i^{\mathrm{o}})
    = \mathrm{LLM}\!\left(
        s_i,\mathcal{S}_i;
        p_{\mathrm{corr}}^{(r_i)},
        p_{\mathrm{o}}^{(d_i)}
      \right), 3 \leq |\widetilde{\mathcal{S}}_i| \leq 5,
\end{equation}
where the Correction Prompt $p_{\mathrm{corr}}^{(r_i)}$ controls one of four
correction structures:
\begin{itemize}
    \item \textbf{No correction:} the speaker makes no revision;
    \item \textbf{Single correction:} the speaker changes $A$ to $B$ once;
    \item \textbf{Rollback correction:} the speaker changes $A$ to $B$ and then
    returns to $A$;
    \item \textbf{Multiple correction:} the speaker changes $A$ to $B$ and later
    revises it to $C$.
\end{itemize}

The Oral Prompt $p_{\mathrm{o}}^{(d_i)}$\allowbreak{}
independently controls the degree of
spoken-style variation, where
$d_i\in\{\mathrm{Low},\mathrm{Moderate},\mathrm{High}\}$.\allowbreak{}
These levels introduce
increasing frequencies and varieties of fillers, repetitions, stuttering, and
other semantically neutral spoken-language phenomena. For the
\emph{explanation} scene, we augment the Correction Prompt with
$p_{\mathrm{exp}}$,\allowbreak{}
which requires the LLM to explain rare, unfamiliar, or
ambiguous nouns.

We designate pass-through cases during Oral generation and ensure that they
constitute 8\% of the final training corpus. For these examples, both control
prompts are disabled, and the utterance contains neither corrections nor
colloquial phenomena. They also bypass ASR Simulation, so the same clean
sentence serves as both the input and target.

\paragraph{Stage 3: Clean generation.}
The Clean Prompt $p_{\mathrm{c}}$ applies the Oral-to-Written transformation to
produce a clean written target by resolving the final intended content and
applying ITN and formatting:
\begin{equation}
    T_i^{\mathrm{c}} = \mathrm{LLM}(T_i^{\mathrm{o}}; p_{\mathrm{c}}).
\end{equation}

\paragraph{Stage 4: ASR simulation.}
After generating the Clean target, we use an LLM with a dedicated ASR Simulation
Prompt $p_{\mathrm{asr}}$ to construct an ASR-style input and its aligned target:
\begin{equation}
    (H_i,Y_i)
    = \mathrm{LLM}(T_i^{\mathrm{o}},T_i^{\mathrm{c}}; p_{\mathrm{asr}}).
\end{equation}
This stage models errors in ASR hypotheses rather than the spoken-language
phenomena already introduced during Oral generation. We apply this
truncation-and-corruption branch to 20\% of generated pairs. For each selected
pair, the prompt samples a truncation boundary and returns a partial hypothesis
$H_i$ together with the correspondingly truncated Clean target $Y_i$. It also
introduces word omissions, recognition errors, and irregular punctuation into
$H_i$. Each $(H_i,Y_i)$ pair is an ordinary text-to-text example: $H_i$ is one
input string and $Y_i$ is one target string. The data contain no chunk slots or
chunk-specific output targets. The aligned pair prevents the Refiner from
predicting content beyond the observed partial utterance while exposing it to
noisy intermediate ASR outputs. The remaining pairs retain their full-length
input--target alignment.

\paragraph{Stage 5: Quality control and near-duplicate filtering.}
Gemma-4-31B-IT first checks whether each Clean target preserves the intended
meaning, resolves self-corrections, omits no intended content, and introduces no
unsupported content. For truncated examples, it also verifies that $H_i$ and
$Y_i$ end at the same semantic boundary and that $Y_i$ contains no content
beyond the observed partial utterance. We then perform global near-duplicate
filtering using text 3-gram Jaccard similarity. For each candidate pair, we
compare its text with the retained pairs and discard the candidate if any
similarity score is at least 0.75. Each remaining pair $(H_i,Y_i)$ is used to
train the Refiner.

% The implementation diagram should depict one concatenated active-context input
% and one refined output string. Do not depict a structured per-chunk model output.

\subsubsection{AgenticASR System}

AgenticASR uses a two-stage ASR--Refiner architecture. The ASR front end
produces a spoken-form hypothesis, and the
LM-based Refiner transforms a text string derived from that hypothesis into
clean written text that preserves final intent. The Refiner is post-trained on
pairs generated by the preceding pipeline and can be shared across ASR front
ends. We evaluate this design with Qwen3-ASR and Whisper.

\paragraph{Offline inference.}
Given a complete audio input $A$, the ASR model first produces the complete
hypothesis $H=F_{\mathrm{ASR}}(A)$. The Refiner then applies the
Oral-to-Written transformation once after the utterance ends, yielding
$Y=F_{\mathrm{R}}(H)$.

\paragraph{Online inference.}
For online processing, an incremental ASR model emits partial hypotheses. Voice
activity detection (VAD) provides speech-boundary signals. A Chunk Manager uses
these signals and sentence-final punctuation to identify stable source-text
spans. To bound the text processed at each update, it limits each span to $L=80$
characters. If the limit is reached, the manager closes the span at the nearest
preceding punctuation or at the limit when none is available.

The chunks are scheduling units, not separate Refiner input or output slots.
At update $t$, the manager selects the current source span $C_t$ and up to
$K-1$ preceding spans, concatenates their ASR text, and sends the resulting
ordinary text string to the Refiner:
\begin{equation}
    \begin{aligned}
        W_t &= C_{\max(1,t-K+1)} \mathbin{\Vert} \cdots \mathbin{\Vert} C_t, \\
        \widehat{Y}_t &= F_{\mathrm{R}}(W_t),
    \end{aligned}
\end{equation}
where $\mathbin{\Vert}$ denotes text concatenation and $\widehat{Y}_t$ is one
clean written string. The system replaces the previously emitted clean-text span
associated with the selected source window by $\widehat{Y}_t$. When a later
source span enters the window, the system repeats this operation on the shifted
window, so new right context can revise an earlier local output.

We use $K=3$ by default. The fixed window bounds the context and computation of
each Refiner call rather than the duration of the audio stream. AgenticASR can
therefore process ongoing audio through repeated bounded updates without passing
the complete stream to the Refiner. Text outside the selected source window is
retained in the emitted transcript. Task-specific post-training makes a compact
Refiner practical for these repeated transformations.

\begin{table*}[htbp!]
    \centering
    \small
    \setlength{\tabcolsep}{1.5pt}
    \begin{tabular}{@{}llrrrrrrr@{}}
        \toprule
        \multicolumn{2}{c}{Configuration}
        & \multirow{2}{*}{Content $\uparrow$}
        & \multirow{2}{*}{Format $\uparrow$}
        & \multirow{2}{*}{Filter $\uparrow$}
        & \multirow{2}{*}{Rephrase $\uparrow$}
        & \multirow{2}{*}{\shortstack{WER/CER/MER (\%) $\downarrow$}}
        & \multirow{2}{*}{Latency (s) $\downarrow$}
        & \multirow{2}{*}{Overall $\uparrow$} \\
        \cmidrule(lr){1-2}
        ASR Model & LM & & & & & & & \\
        \midrule
        Qwen3-ASR-0.6B & Qwen3.5-Flash
            & 87.50 & 28.97 & 73.13 & 49.13
            & 26.82/17.01/21.91 & 60.08 & 66.47 \\
        FormalASR-0.6B & --
            & 86.63 & 14.35 & 36.51 & 13.29
            & 38.67/28.34/34.38 & \textbf{3.42} & 48.76 \\
        Qwen3-ASR-0.6B & AgenticASR
            & 87.30 & 54.94 & 78.80 & 69.16
            & 14.64/7.79/10.23 & 6.60 & 76.15 \\
        \addlinespace[2pt]
        Qwen3-ASR-1.7B & Qwen3.5-Flash
            & 90.21 & 35.48 & 75.82 & 52.10
            & 24.60/15.72/20.29 & 60.89 & 69.93 \\
        FormalASR-1.7B & --
            & 90.11 & 19.69 & 40.59 & 15.70
            & 34.07/24.47/30.48 & 3.46 & 52.50 \\
        Qwen3-ASR-1.7B & AgenticASR
            & \textbf{90.24} & \textbf{65.19}
            & \textbf{78.89} & \textbf{72.83}
            & \textbf{12.70/6.86/9.01} & 9.59 & \textbf{79.95} \\
        \midrule
        Whisper Base & Gemini-2.5-Flash
            & 47.04 & 6.09 & 62.63 & 16.67
            & 53.14/39.83/46.41 & 12.00 & 37.09 \\
        Whisper Base & AgenticASR
            & 38.69 & 6.95 & 71.96 & 32.47
            & 55.62/41.33/46.70 & 5.86 & 38.82 \\
        Whisper Small & Gemini-2.5-Flash
            & 58.79 & 29.04 & 65.08 & 27.94
            & 56.98/43.74/45.78 & 11.99 & 48.78 \\
        Whisper Small & AgenticASR
            & 52.58 & 29.57 & 72.79 & 47.40
            & 58.09/44.09/44.28 & 6.89 & 51.72 \\
        Whisper Large & Gemini-2.5-Flash
            & \textbf{80.23} & 51.58 & 63.10 & 36.13
            & 31.50/21.45/25.33 & 8.04 & 62.90 \\
        Whisper Large & AgenticASR
            & 76.16 & \textbf{55.87} & \textbf{77.75}
            & \textbf{63.01} & \textbf{27.51/18.19/19.63}
            & \textbf{4.42} & \textbf{70.29} \\
        \bottomrule
    \end{tabular}
    \caption{Main results on AASR-Bench. WER, CER, and MER are token-level
    metrics, and latency is the mean end-to-end inference time. The LM column
    identifies the downstream transformation system; FormalASR performs direct
    speech-to-clean-text recognition and therefore has no separate LM. Best
    values within each ASR family are shown in bold.}
    \label{tab:main-results}
\end{table*}

\section{Experiments}

\subsection{Experimental Setup}

Using the data generation pipeline described above, we constructed 100,000
input--target pairs, using 85\% for training and 15\% for validation to select
the default configuration and monitor overfitting. The Refiner was
initialized from MiniCPM-5-1B \cite{minicpmteam2025minicpm4}, and
Qwen3-ASR-1.7B served as the default ASR front end \cite{shi2026qwen3}. We used
full-parameter supervised fine-tuning for five epochs with AdamW, a learning
rate of $2\times10^{-5}$, cosine decay, a 5\% warmup ratio, 0.01 weight decay,
and gradient clipping at 1.0. The per-device batch size was 16 with two gradient
accumulation steps. On four NVIDIA H100 GPUs, training used BF16, gradient
checkpointing, packed sequences, and a maximum sequence length of 1,024 tokens,
with loss applied only to target tokens.

WER, CER, and MER were computed after text normalization for English, Chinese,
and mixed Chinese--English utterances, respectively. Latency denotes the observed
end-to-end time from input to final output. All inference-latency measurements
were obtained on CPUs with CUDA disabled. For all LLM-judge operations, each
rubric question was judged three times, and the final option was chosen by
majority vote to reduce judgment variance.

\subsection{Main Results}

\begin{figure}[H]
    \centering
    \includegraphics[width=0.8\columnwidth]{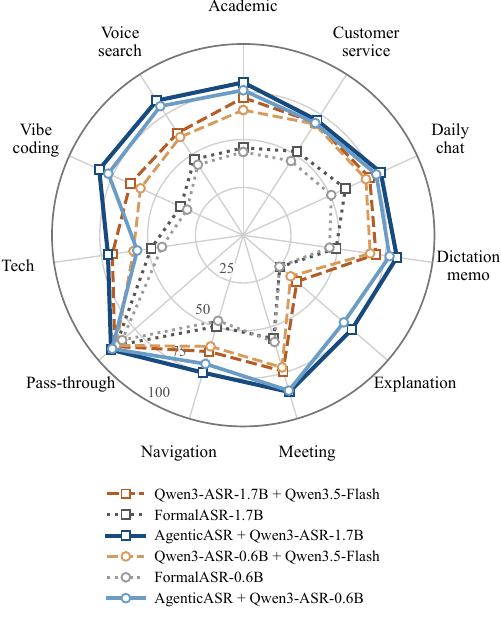}
    \caption{Scene-level Overall scores for the Qwen3-ASR family across ten
    usage scenes and the pass-through control. Solid lines denote AgenticASR
    with the MiniCPM-5-1B Refiner, dashed lines denote Qwen3.5-Flash API-based
    transformation, and dotted lines denote FormalASR.}
    \label{fig:qwen-scene-radar}
\end{figure}

Table~\ref{tab:main-results} groups the Qwen systems by ASR scale and places
AgenticASR after the two corresponding baselines. AgenticASR achieved the highest
Overall score among the evaluated systems on AASR-Bench. With the default
MiniCPM-5-1B Refiner and Qwen3-ASR-1.7B, AgenticASR scored 79.95 Overall and led
all four rubric dimensions. Across the five matched ASR backbones, AgenticASR's
Overall advantage over
API-based transformation ranged from 1.73 to 10.02 points, with substantially
lower latency. Although slower than FormalASR, AgenticASR improved Overall by
approximately 27.4 points at both Qwen scales. The Qwen configurations also
achieved low token-level error rates. AASR-Bench further exposes differences in
formatting, filtering, and correction resolution that WER, CER, and MER do not
capture.

The Whisper results further expose dependence on the upstream ASR model.
AgenticASR improved Overall at every Whisper scale. The advantage over the API
baseline widened from 1.73 points with Base to 7.39 points with Large. The
gains came primarily from Filter and Rephrase, whereas Content remained below the
API baseline at all three scales. With Whisper Base, token-level error rates also
remained close to those of the API baseline, consistent with limited semantic
evidence for downstream transformation. A general-purpose API such as Gemini may
use stronger zero-shot rewriting to reorganize an incomplete transcript, whereas
our compact Refiner is optimized for evidence-supported Oral-to-Written
transformation rather than unconstrained content reconstruction. The stronger
results with Whisper Large indicate that transformation quality remains closely
coupled to the evidence preserved by the upstream ASR model.

Figure~\ref{fig:qwen-scene-radar} shows that AgenticASR with Qwen3-ASR-1.7B led
both baselines in all 10 usage scenes and the pass-through control. With the
0.6B front end, it led in nine scenes and the control; the exception was Tech,
where the API baseline was 1.98 points higher. The largest AgenticASR advantages
occurred in Explanation, Voice search, and Vibe coding, where resolving revisions
and specialized entities is especially important. The 1.7B backbone improved
the full scene profile, consistent with stronger ASR models preserving better
evidence for transformation. Notably, both FormalASR variants scored lower on
Pass-through than the API and AgenticASR systems at the corresponding scale.
This result suggests that directly
post-training an ASR model for correction can bias it toward rewriting already
clean utterances, thereby weakening transcription faithfulness and increasing
the risk of hallucinated edits.
\enlargethispage{\baselineskip}

\subsection{Ablation Study}

\paragraph{Human Agreement with the Rubric-Based Judge.}
To assess whether rubric-based scores reflect human judgment, we sampled 10
utterances from each usage scene, yielding 100 utterances after excluding the
pass-through control. Chinese samples exceeded 10 characters, and English
samples exceeded 10 words. Two domain experts independently answered
each applicable rubric for AgenticASR outputs from the Qwen3-ASR-0.6B and
Qwen3-ASR-1.7B front ends. The evaluation was double-blind: annotators saw
anonymized outputs in randomized order, were not told the generating system, and
did not see either the automatic judgment or the other annotator's answers. For
each ASR front end and each annotator, we compared human and Gemma-4-31B-IT
answers at the rubric-question level. Table~\ref{tab:human-judge-alignment}
reports the mean of the two annotator-level Spearman correlations
\cite{spearman1904} and quadratic-weighted kappa values
\cite{cohen1968,fleiss1973}. These results indicate strong agreement between
Gemma-4-31B-IT and independent human assessments in this validation sample.

\begin{table}[htbp!]
    \centering
    \small
    \setlength{\tabcolsep}{5pt}
    \begin{tabular}{@{}lcc@{}}
        \toprule
        & \shortstack{Qwen3-ASR-\\0.6B} & \shortstack{Qwen3-ASR-\\1.7B} \\
        \midrule
        Spearman $\rho$ & 0.8222 & 0.8064 \\
        Quadratic-weighted $\kappa$ & 0.8313 & 0.7918 \\
        \bottomrule
    \end{tabular}
    \caption{Mean answer-level agreement between the rubric-based Gemma-4-31B-IT
    judge and two independent domain experts under double-blind evaluation. Higher is
    better.}
    \label{tab:human-judge-alignment}
\end{table}

\paragraph{Effect of Refiner Capacity.}
To isolate the effect of Refiner capacity, we fixed the ASR front end to
Qwen3-ASR-1.7B and compared three Refiner sizes.
Table~\ref{tab:refiner-size-ablation} reports the aggregate results.

\begin{table}[htbp!]
    \centering
    \small
    \setlength{\tabcolsep}{2.2pt}
    \begin{tabular}{@{}lrrrrrr@{}}
        \toprule
        Refiner & Overall & Cont. & Fmt. & Filt. & Reph. & Lat. (s) \\
        \midrule
        Qwen2.5-0.5B-Instruct             & 78.76 & 88.00 & 63.40 & 78.36 & 69.85 & 9.21 \\
        MiniCPM-5-1B                      & 79.95 & 90.24 & 65.19 & 78.89 & 72.83 & 9.59 \\
        Qwen2.5-4B-Instruct               & \textbf{83.42} & \textbf{91.00}
                                  & \textbf{74.43} & \textbf{83.31}
                                  & \textbf{75.68} & 10.77 \\
        \bottomrule
    \end{tabular}
    \caption{Refiner capacity comparison with Qwen3-ASR-1.7B as the fixed front end. Best results are boldfaced.}
    \label{tab:refiner-size-ablation}
\end{table}

Overall increased monotonically with Refiner size: the 4B model exceeded the
0.5B model by 4.66 points. The largest gains occurred in Format (+11.03) and
Rephrase (+5.83), indicating that larger Refiners' stronger semantic
understanding improves contextual formatting and final-intent resolution;
Content also improved by 3.00 points. Latency rose from 9.21~s to 10.77~s.
Thus, larger Refiners suit latency-tolerant offline use, whereas smaller ones
better meet online latency constraints.

\paragraph{Ablation of Online Inference.}
Offline AgenticASR feeds the complete Qwen3-ASR-1.7B transcript to the
Refiner in a single pass. Online AgenticASR instead refines a sliding window
of $K$ consecutive source chunks. Table~\ref{tab:online-comparison} reports
Rephrase, Explanation, and end-to-end latency.

\begin{table}[htbp!]
    \centering
    \small
    \setlength{\tabcolsep}{2.5pt}
    \begin{tabular}{@{}lrrr@{}}
        \toprule
        Setting & Rephrase $\uparrow$ & Latency (s) $\downarrow$
        & Explanation $\uparrow$ \\
        \midrule
        Offline      & 72.83 &  9.59 & 75.20 \\
        Window $=1$  & 36.17 & 11.28 & 19.43 \\
        Window $=2$  & 65.08 & 11.70 & 55.06 \\
        Window $=3$  & 70.47 & 12.15 & 74.00 \\
        \bottomrule
    \end{tabular}
    \caption{Offline and online AgenticASR with Qwen3-ASR-1.7B. Rephrase is a
    rubric-dimension score, Explanation is the scene-level Overall score, and
    latency is the mean end-to-end inference time.}
    \label{tab:online-comparison}
\end{table}

Larger windows sharply improve quality with only a modest latency increase.
Moving from $K{=}1$ to $K{=}3$ raises Rephrase from 36.17 to 70.47 (offline:
72.83) and Explanation from 19.43 to 74.00 (offline: 75.20), while latency
grows by just 0.87\,s (11.28 to 12.15).

The jump from $K{=}1$ to $K{=}2$ ($+28.91$ Rephrase, $+35.63$ Explanation)
arises from cross-chunk self-repairs. When VAD places a boundary inside a
correction such as \textit{``I want to go to Beijing <VAD> oh no, Shanghai,''} a
single-chunk window processes the two fragments independently and cannot
retroactively revise the already emitted ``Beijing''. With $K{=}2$, the prior
chunk is concatenated before refinement, so the model recognizes the repair
pair and correctly outputs ``Shanghai''. The further gain from $K{=}2$ to
$K{=}3$ is especially large for Explanation ($+18.94$), consistent with
spelling and explanatory cues being distributed over longer local spans. At
$K{=}3$, the gaps to offline scores shrink to 2.36 (Rephrase) and 1.20
(Explanation). Hence, a three-chunk window recovers nearly all useful right
context on AASR-Bench while adding under one second of end-to-end latency.

\section{Conclusion}

This work advances speech recognition from literal transcripts toward usable
written text. We defined AgenticSR, constructed AASR-Bench for bilingual
atomic-rubric evaluation, and developed AgenticASR with an LLM-assisted pipeline
for training its compact Oral-to-Written Refiner. The Refiner can serve ASR
front ends and revise a bounded span as speech arrives. Across ASR backbones and
Refiner scales, AgenticASR outperformed the evaluated baselines on AASR-Bench.
Human agreement supported the
rubric-based evaluation, and a three-chunk window approached offline quality
with modest additional latency. Performance nevertheless depends on the
evidence retained by the upstream ASR front end, making recognition quality an
important deployment constraint.

Looking ahead, AgenticASR could support voice assistants, meeting
transcription, live dictation, and other streaming speech interfaces that
require readable text before an utterance is complete. Its local replacement
mechanism can incorporate later corrections, spellings, or explanations without
revisiting an unbounded stream. Combining this mechanism with stronger
multilingual front ends and broader conversational evaluation could make
intent-preserving transcription practical across a wider range of real-time
speech interactions.

\appendix

This appendix provides the detailed evidence underlying the aggregate results
and evaluation protocol in the main text. Section~\ref{sec:detailed-results}
breaks AASR-Bench performance down by usage scene and rubric dimension for each
ASR front end, and also reports the Refiner-capacity comparison. Section~\ref{sec:online-case}
illustrates why online revision requires more than one source chunk. Finally,
Section~\ref{sec:rubric-example} gives a worked example of the atomic rubrics used
to score Content preservation, written-form Format, Filter operations, and
Rephrase operations.

\section{Detailed AASR-Bench Results}
\label{sec:detailed-results}

The following tables expand the main results from aggregate scores to individual
usage scenes. All systems were evaluated on the full AASR-Bench test set with
the question-weighted scoring formula defined in the main paper; higher values
are better. In each table, the first block reports the Overall score within each
scene and on the pass-through control. The second block aggregates rubric
questions by dimension, and the final row gives the Overall score across the
complete benchmark. Content applies to all samples, whereas Format, Filter, and
Rephrase are scored only when the corresponding phenomenon is present.

\subsection{Qwen3-ASR Front Ends}

Tables~\ref{tab:supp-qwen-06} and~\ref{tab:supp-qwen-17} compare three
transformation strategies while holding the Qwen3-ASR front end fixed:
AgenticASR uses the compact MiniCPM-5-1B Refiner, FormalASR directly adapts the
ASR model, and OpenTypeless uses Qwen3.5-Flash for API-based post-transformation.
This matched-front-end design isolates differences in the Oral-to-Written
transformation rather than differences in the initial speech recognizer.

\begin{table}[htbp]
    \centering
    \small
    \setlength{\tabcolsep}{6pt}
    \begin{tabular}{@{}lrrr@{}}
        \toprule
        Metric & AgenticASR & FormalASR & OpenTypeless \\
        \midrule
        Academic & 75.80 & 43.53 & 65.44 \\
        Customer service & 69.70 & 46.15 & 69.09 \\
        Daily chat & 76.49 & 50.62 & 70.53 \\
        Dictation memo & 77.29 & 45.61 & 67.08 \\
        Explanation & 69.60 & 25.07 & 32.80 \\
        Meeting & 84.68 & 58.21 & 72.06 \\
        Navigation & 70.11 & 46.65 & 60.61 \\
        Tech & 56.01 & 42.91 & 57.99 \\
        Vibe coding & 77.56 & 32.24 & 59.04 \\
        Voice search & 80.19 & 43.82 & 60.98 \\
        Pass-through & 90.68 & 83.73 & 87.93 \\
        \midrule
        Content & 87.30 & 86.63 & 87.50 \\
        Format & 54.94 & 14.35 & 28.97 \\
        Filter & 78.80 & 36.51 & 73.13 \\
        Rephrase & 69.16 & 13.29 & 49.13 \\
        \midrule
        Overall & 76.15 & 48.76 & 66.47 \\
        \bottomrule
    \end{tabular}
    \caption{Detailed comparison of AgenticASR, FormalASR, and OpenTypeless
    with Qwen3-ASR-0.6B.}
    \label{tab:supp-qwen-06}
\end{table}

With the 0.6B front end (Table~\ref{tab:supp-qwen-06}), AgenticASR reached
76.15 Overall, exceeding OpenTypeless by 9.68 points and FormalASR by 27.39
points. The gains were concentrated in transformation-sensitive dimensions:
AgenticASR improved Format from 28.97 to 54.94 and Rephrase from 49.13 to 69.16
relative to OpenTypeless, while maintaining comparable Content preservation.
At the scene level, the largest advantage over OpenTypeless occurred for
Explanation (69.60 versus 32.80); Tech was the only scene in which OpenTypeless
scored higher (57.99 versus 56.01).

\begin{table}[htbp]
    \centering
    \small
    \setlength{\tabcolsep}{6pt}
    \begin{tabular}{@{}lrrr@{}}
        \toprule
        Metric & AgenticASR & FormalASR & OpenTypeless \\
        \midrule
        Academic & 79.88 & 45.72 & 71.91 \\
        Customer service & 71.42 & 52.01 & 69.91 \\
        Daily chat & 79.06 & 58.73 & 72.69 \\
        Dictation memo & 81.20 & 49.05 & 70.13 \\
        Explanation & 75.20 & 25.33 & 37.07 \\
        Meeting & 85.42 & 56.37 & 74.14 \\
        Navigation & 74.86 & 49.72 & 63.41 \\
        Tech & 71.45 & 48.65 & 69.12 \\
        Vibe coding & 82.79 & 35.95 & 64.92 \\
        Voice search & 83.67 & 47.18 & 63.51 \\
        Pass-through & 91.47 & 88.06 & 88.85 \\
        \midrule
        Content & 90.24 & 90.11 & 90.21 \\
        Format & 65.19 & 19.69 & 35.48 \\
        Filter & 78.89 & 40.59 & 75.82 \\
        Rephrase & 72.83 & 15.70 & 52.10 \\
        \midrule
        Overall & 79.95 & 52.50 & 69.93 \\
        \bottomrule
    \end{tabular}
    \caption{Detailed comparison of AgenticASR, FormalASR, and OpenTypeless
    with Qwen3-ASR-1.7B.}
    \label{tab:supp-qwen-17}
\end{table}

With the stronger 1.7B front end (Table~\ref{tab:supp-qwen-17}), all three
systems improved, but their ordering remained unchanged. AgenticASR achieved
79.95 Overall, 10.02 points above OpenTypeless and 27.45 points above FormalASR.
Content scores were nearly identical across systems, so the separation mainly
came from Format, Filter, and Rephrase. AgenticASR also led OpenTypeless in all
ten usage scenes and on the pass-through control, indicating that the gain was
not confined to one domain.

\subsection{Whisper Front Ends}

Tables~\ref{tab:supp-whisper-base}--\ref{tab:supp-whisper-small} compare
AgenticASR with Gemini-2.5-Flash-based OpenTypeless while holding each Whisper
front end fixed. These results test whether the same Refiner remains useful when
the upstream ASR model supplies weaker or stronger lexical and semantic
evidence. They also expose the central boundary of the two-stage design: a
Refiner can remove spoken artifacts and resolve corrections, but it cannot
reliably recover content that the ASR front end failed to preserve.

\begin{table}[htbp]
    \centering
    \small
    \setlength{\tabcolsep}{10pt}
    \begin{tabular}{@{}lrr@{}}
        \toprule
        Metric & AgenticASR & OpenTypeless \\
        \midrule
        Academic & 34.86 & 33.85 \\
        Customer service & 45.80 & 43.26 \\
        Daily chat & 26.59 & 26.96 \\
        Dictation memo & 38.93 & 38.89 \\
        Explanation & 33.87 & 21.78 \\
        Meeting & 35.78 & 31.99 \\
        Navigation & 45.81 & 34.36 \\
        Tech & 30.15 & 32.35 \\
        Vibe coding & 35.29 & 28.32 \\
        Voice search & 48.98 & 44.50 \\
        Pass-through & 48.69 & 56.80 \\
        \midrule
        Content & 38.69 & 47.04 \\
        Format & 6.95 & 6.09 \\
        Filter & 71.96 & 62.63 \\
        Rephrase & 32.47 & 16.67 \\
        \midrule
        Overall & 38.82 & 37.09 \\
        \bottomrule
    \end{tabular}
    \caption{Detailed comparison of AgenticASR and OpenTypeless with Whisper
    Base.}
    \label{tab:supp-whisper-base}
\end{table}

For Whisper Base (Table~\ref{tab:supp-whisper-base}), AgenticASR improved
Overall by 1.73 points. It produced clear gains in Filter (71.96 versus 62.63)
and Rephrase (32.47 versus 16.67), but its Content score was 8.35 points lower.
The lower Content and pass-through scores show that transformation gains cannot
fully offset recognition errors from the smallest Whisper front end.

\begin{table}[htbp]
    \centering
    \small
    \setlength{\tabcolsep}{10pt}
    \begin{tabular}{@{}lrr@{}}
        \toprule
        Metric & AgenticASR & OpenTypeless \\
        \midrule
        Academic & 68.13 & 63.52 \\
        Customer service & 69.44 & 64.47 \\
        Daily chat & 66.22 & 59.96 \\
        Dictation memo & 70.42 & 63.93 \\
        Explanation & 56.27 & 33.33 \\
        Meeting & 80.02 & 66.18 \\
        Navigation & 67.24 & 51.44 \\
        Tech & 56.55 & 53.86 \\
        Vibe coding & 74.95 & 61.44 \\
        Voice search & 73.71 & 64.95 \\
        Pass-through & 80.84 & 83.60 \\
        \midrule
        Content & 76.16 & 80.23 \\
        Format & 55.87 & 51.58 \\
        Filter & 77.75 & 63.10 \\
        Rephrase & 63.01 & 36.13 \\
        \midrule
        Overall & 70.29 & 62.90 \\
        \bottomrule
    \end{tabular}
    \caption{Detailed comparison of AgenticASR and OpenTypeless with Whisper
    Large.}
    \label{tab:supp-whisper-large}
\end{table}

Whisper Large supplied substantially stronger source hypotheses
(Table~\ref{tab:supp-whisper-large}). Under this front end, AgenticASR improved
Overall from 62.90 to 70.29 and led OpenTypeless in every transformed usage
scene. The largest dimension-level gains were in Rephrase (26.88 points) and
Filter (14.65 points). Content remained 4.07 points lower and pass-through was
2.76 points lower, which makes the remaining preservation trade-off explicit.

\begin{table}[htbp]
    \centering
    \small
    \setlength{\tabcolsep}{10pt}
    \begin{tabular}{@{}lrr@{}}
        \toprule
        Metric & AgenticASR & OpenTypeless \\
        \midrule
        Academic & 46.22 & 44.52 \\
        Customer service & 52.63 & 53.77 \\
        Daily chat & 46.41 & 41.79 \\
        Dictation memo & 45.23 & 48.09 \\
        Explanation & 44.27 & 29.33 \\
        Meeting & 62.13 & 53.68 \\
        Navigation & 55.03 & 42.18 \\
        Tech & 35.01 & 35.01 \\
        Vibe coding & 58.82 & 53.59 \\
        Voice search & 59.54 & 51.62 \\
        Pass-through & 63.65 & 68.24 \\
        \midrule
        Content & 52.58 & 58.79 \\
        Format & 29.57 & 29.04 \\
        Filter & 72.79 & 65.08 \\
        Rephrase & 47.40 & 27.94 \\
        \midrule
        Overall & 51.72 & 48.78 \\
        \bottomrule
    \end{tabular}
    \caption{Detailed comparison of AgenticASR and OpenTypeless with Whisper
    Small.}
    \label{tab:supp-whisper-small}
\end{table}

Whisper Small showed the same intermediate pattern
(Table~\ref{tab:supp-whisper-small}). AgenticASR raised Overall by 2.94 points,
including gains of 19.46 points in Rephrase and 7.71 points in Filter, while
Content decreased by 6.21 points. Scene-level gains were broad but not uniform:
OpenTypeless remained higher for Customer service, Dictation memo, and the
pass-through control, and the two systems tied on Tech.

\subsection{Effect of Refiner Capacity}

The final detailed table holds Qwen3-ASR-1.7B fixed and changes only the Refiner.
It therefore separates Refiner capacity from front-end ASR quality. The scene
rows show where additional capacity changes performance, while the rubric rows
identify which transformation operations account for the aggregate difference.

\begin{table}[htbp]
    \centering
    \small
    \setlength{\tabcolsep}{4pt}
    \begin{tabular}{@{}lrrr@{}}
        \toprule
        Metric & \shortstack{Qwen2.5-0.5B\\Instruct} & \shortstack{MiniCPM-5-1B} & \shortstack{Qwen2.5-4B\\Instruct} \\
        \midrule
        Academic & 76.89 & 79.88 & 82.97 \\
        Customer service & 72.65 & 71.42 & 81.95 \\
        Daily chat & 79.06 & 79.06 & 82.44 \\
        Dictation memo & 81.39 & 81.20 & 84.92 \\
        Explanation & 78.40 & 75.20 & 80.27 \\
        Meeting & 87.75 & 85.42 & 87.50 \\
        Navigation & 70.67 & 74.86 & 74.58 \\
        Tech & 70.02 & 71.45 & 73.43 \\
        Vibe coding & 81.48 & 82.79 & 82.79 \\
        Voice search & 82.59 & 83.67 & 84.99 \\
        Pass-through & 92.13 & 91.47 & 92.65 \\
        \midrule
        Content & 88.00 & 90.24 & 91.00 \\
        Format & 63.40 & 65.19 & 74.43 \\
        Filter & 78.36 & 78.89 & 83.31 \\
        Rephrase & 69.85 & 72.83 & 75.68 \\
        \midrule
        Overall & 78.76 & 79.95 & 83.42 \\
        \bottomrule
    \end{tabular}
    \caption{Effect of Refiner capacity with Qwen3-ASR-1.7B fixed as the
    AgenticASR front end.}
    \label{tab:supp-refiner-capacity}
\end{table}

Table~\ref{tab:supp-refiner-capacity} shows a monotonic increase in Overall
score, from 78.76 with the 0.5B Refiner to 79.95 with the 1B Refiner and 83.42
with the 4B Refiner. Relative to the 0.5B model, the 4B model gained 11.03 points
in Format, 5.83 points in Rephrase, 4.95 points in Filter, and 3.00 points in
Content. These results indicate that additional Refiner capacity primarily
benefits structured rewriting operations, although the latency results in the
main paper show that this quality gain must be balanced against response time.

\section{Case Study on Online Inference}
\label{sec:online-case}

Online AgenticASR treats voice-activity-detection (VAD) chunks as scheduling
units. At step $t$, the Chunk Manager concatenates the current chunk with up to
$K-1$ preceding source chunks, sends this ordinary text string to the Refiner,
and replaces the output span associated with that active window. Increasing $K$
therefore gives the Refiner more right context with which to revise earlier
content; it does not create separate model inputs or outputs for each chunk.

\begin{figure}[htbp]
    \centering
    \includegraphics[width=\linewidth]{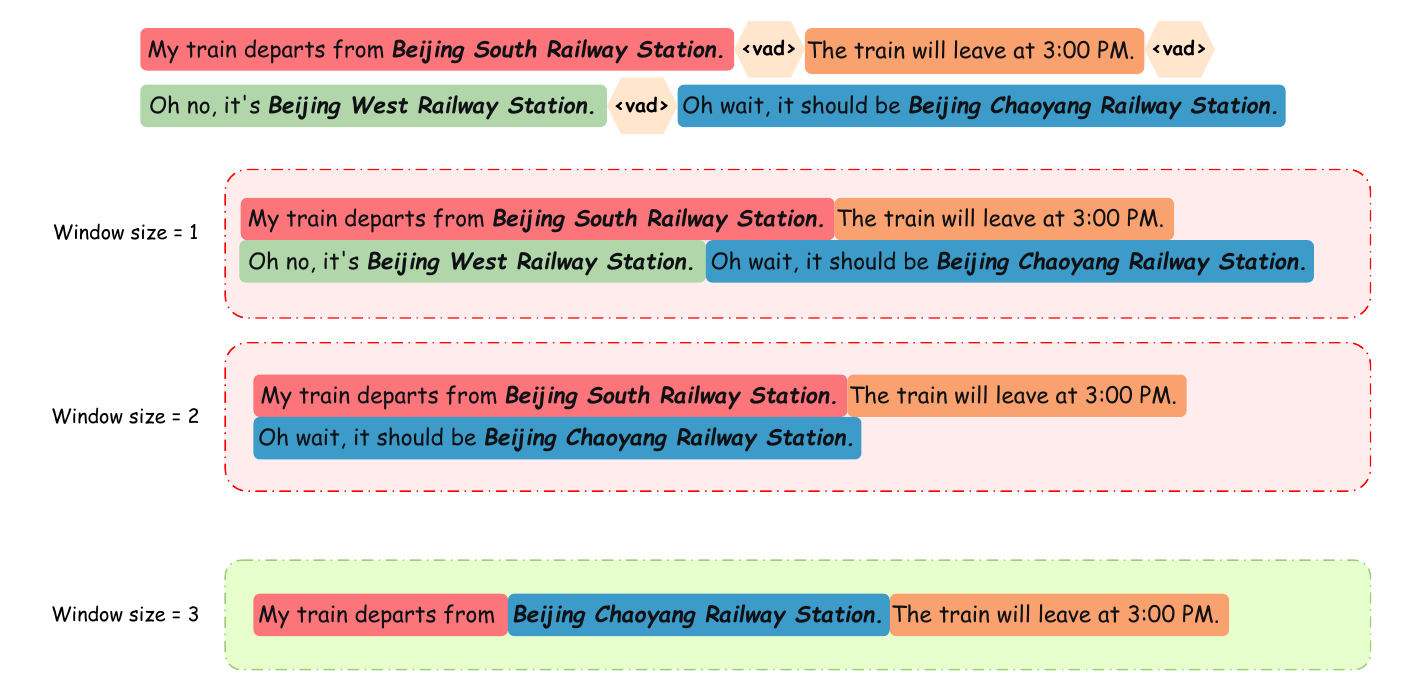}
    \caption{Effect of active-window size on online revision. The utterance
    first names Beijing South Railway Station, then revises the destination to
    Beijing West Railway Station and finally to Beijing Chaoyang Railway
    Station. With window size $1$, the system cannot connect the corrections
    across VAD boundaries. Window size $2$ removes the immediately preceding
    alternative but leaves the earlier destination. Window size $3$ retains
    sufficient local context to produce the final intended destination while
    preserving the departure time.}
    \label{fig:supp-window-context}
\end{figure}

Figure~\ref{fig:supp-window-context} makes this mechanism concrete with a
multi-stage destination correction that crosses VAD boundaries. With a
one-chunk window, the Refiner sees each fragment independently, so both
superseded destination names remain in the transcript. A two-chunk window joins
the final correction to the immediately preceding alternative and removes that
alternative, but the earliest destination lies outside the active context. A
three-chunk window covers the full repair sequence, allowing the system to retain
the departure time while replacing all abandoned destinations with the final
intended one. This example explains the quantitative trend in the main paper:
$K=3$ nearly matches offline Rephrase and Explanation scores because it captures
corrections and explanatory cues distributed across multiple local spans.

\section{Worked Example of Rubric-Based Evaluation}
\label{sec:rubric-example}

AASR-Bench uses atomic questions because a single transcript may require several
different operations at once. Table~\ref{tab:rubric_example} starts from an Oral
utterance containing a filler, a repeated phrase, a multi-stage numerical
correction, and a number that requires written-form normalization. The Clean
reference preserves the intended shopping preference and the final price while
removing the abandoned alternatives and rendering the amount as ``\$30.''

The lower part of the table separates these requirements into four independently
interpretable dimensions. Content asks whether an unaffected semantic unit is
preserved. Format asks whether the final amount is normalized appropriately.
Filter checks whether fillers and repetitions are removed, and Rephrase checks
whether the complete correction process is resolved to the final intended
value. A system output is matched to one option for each applicable question;
the option scores follow the scales defined in the main paper. Each question is
judged three times, with majority voting used to select the final option. Thus,
an output can receive credit for a successful operation while still being
penalized for a distinct preservation or transformation error.

\begin{table*}[htbp!]
\centering

\renewcommand{\arraystretch}{1.4}

\begin{tabular}{
>{\centering\arraybackslash}m{1.4cm}
@{\hspace{0.2cm}}
>{\raggedright\arraybackslash}m{3.8cm}
>{\centering\arraybackslash}m{0.55cm}
>{\raggedright\arraybackslash}m{4.7cm}
>{\raggedright\arraybackslash}m{4.6cm}
}

\toprule

% ================= Example pair =================

\textbf{Oral} &
\multicolumn{4}{m{13.3cm}}{
I mean I am keen ... keen on shopping at Aldi.
Cost is... cost is ten dollars... no twenty... wait wrong,
it is thirty dollars for basic stuff.
}
\\[0.5em]

\textbf{Clean} &
\multicolumn{4}{m{13.3cm}}{
I am keen on shopping at Aldi. Cost is \$30 for basic stuff.
}
\\

\midrule

% ================= Rubric table =================

\multicolumn{1}{c}{\textbf{Dimension}} &
\multicolumn{1}{c}{\textbf{Question}} &
\multicolumn{1}{c}{\textbf{Score}} &
\multicolumn{1}{c}{\textbf{Scoring rule}} &
\multicolumn{1}{c}{\textbf{Example output}}
\\

\midrule

% -------- Content --------

\multirow[c]{3}{*}{Content}
&
\multirow[c]{3}{*}{
\parbox[c]{3.8cm}{
\raggedright
Whether ``shopping at Aldi'' is preserved.
}
}
& 1 &
\rule{0pt}{2.2ex}Preserved: the phrase appears with the same meaning. &
``I am keen on shopping at Aldi.''
\\

& & 0 &
\rule{0pt}{2.2ex}Missing or incomplete: the phrase is absent or only partly preserved. &
``I am keen on shopping at Adi.''
\\

& & -1 &
\rule{0pt}{2.2ex}Unrelated or opposite: the output contradicts the intended meaning. &
``I am keen on walking at Adee.''
\\

\midrule

% -------- Format --------

\multirow[c]{3}{*}{Format}
&
\multirow[c]{3}{*}{
\parbox[c]{3.8cm}{
\raggedright
Whether ``thirty dollars'' is normalized to ``\$30''.
}
}
& 1 &
\rule{0pt}{2.2ex}Correctly normalized: ``\$30'' or an equivalent form is used. &
``Cost is \$30.''
\\

& & 0 &
\rule{0pt}{2.2ex}Not normalized: ``thirty dollars'' remains unchanged. &
``Cost is thirty dollars.''
\\

& & -1 &
\rule{0pt}{2.2ex}Missing or incorrect: neither the correct normalized form nor an equivalent appears. &
``Cost is \$20.''
\\

\midrule

% -------- Filter --------

\multirow[c]{2}{*}{Filter}
&
\multirow[c]{2}{*}{
\parbox[c]{3.8cm}{
\raggedright
Whether fillers and repetitions
(e.g., ``I mean'', ``cost is... cost is'')
are removed.
}
}

& 2 &
\rule{0pt}{2.5ex}Fully removed: no fillers or repetitions remain. &
``I am keen on shopping at Aldi. Cost is \$30.''
\\

& & 0 &
\rule{0pt}{2.5ex}Partially retained: some fillers or repetitions remain. &
``I mean, I am keen on shopping at Aldi. Cost is \$30.''
\\

\midrule

% -------- Rephrase --------

\multirow[c]{4}{*}{Rephrase}
&
\multirow[c]{4}{*}{
\parbox[c]{3.8cm}{
\raggedright
Whether ``ten dollars... no twenty... wait wrong,
it is thirty dollars'' is revised to ``thirty dollars''.
}
}
& 2 &
\rule{0pt}{2.2ex}Correctly revised: the final intended result is retained. &
``Cost is \$30 for basic stuff.''
\\

& & 1 &
\rule{0pt}{2.2ex}Partially revised: the revision process is removed, but the final result is missing. &
``Cost is \$20 for basic stuff.''
\\

& & 0 &
\rule{0pt}{2.2ex}Not revised: the original correction process remains. &
``Cost is \$10, no twenty, wait, \$30.''
\\

& & -1 &
\rule{0pt}{2.2ex}Missing or unrelated: the relevant content is absent. &
``Cost is at Aldi.''
\\

\bottomrule

\end{tabular}
\caption{Worked example of rubric-based evaluation. The Oral utterance and its
Clean reference are shown above four atomic questions. Each question isolates
one requirement of the Oral-to-Written transformation and assigns an option
score according to the degree of preservation or correction.}
\label{tab:rubric_example}

\end{table*}

\begin{CJK}{UTF8}{gbsn}

\begin{table*}[t]
\centering

% \caption{Examples from different scenarios in AASR-Bench.
% Each example contains the clean transcript, oral speech input,
% and the refined output generated by AgenticASR.}

\label{tab:aasr_examples}

\renewcommand{\arraystretch}{1.25}

\small

\begin{tabularx}{0.96\linewidth}{
>{\bfseries}p{2.5cm}
X
}

\toprule

% ================= Vibe Coding =================

\midrule
\multicolumn{2}{l}{\textbf{Metric: Vibe Coding}}\\
\midrule

Clean Transcript &
如果 timeout\_ms 设成5000毫秒的话，那我们就能排除掉网络抖动问题。要是后端用 Haskell 写的，那逻辑层可能就卡在内存回收那块吧。
\\

Oral Input &
嗯，如果 timeout ms 设成五千毫秒的话，那我们就能排除掉，排除掉网络抖动问题，我想想啊，要是后端用 Scala 写的…不对，应该是用 Haskell 写的，那逻辑层可能哈，可能就卡在内存回收那块吧。
\\

AgenticASR Output &
如果timeout ms设置成5000毫秒的话，那我们就能排除掉网络抖动问题。要是后端用Haskell写的，那逻辑层可能就卡在内存回收那块儿吧。
\\

% ================= Explanation =================

\midrule
\multicolumn{2}{l}{\textbf{Metric: Explanation}}\\
\midrule

Clean Transcript &
周六上午先去栖霞寺，下午再爬崂山。
\\

Oral Input &
周六上午先去七霞寺...不对，栖霞寺，栖是栖息的栖，下午再爬牢山，不是，崂山，崂是山字旁一个劳累的劳。
\\

AgenticASR Output &
周六上午先去栖霞寺，下午再爬崂山。
\\

% ================= Academic =================
\midrule
\multicolumn{2}{l}{\textbf{Metric: Academic}}\\
\midrule

Clean Transcript &
So we started by using contrastive learning for the initial representation,
but then we shifted to a curriculum learning approach to stabilize training.
We're seeing perplexity drop to about 4.2, which is great, but we're still
trying to optimize the knowledge distillation process from PaLM 2.
We used a teacher model with 544 billion parameters and we're compressing it
into a smaller student, right?
\\

Oral Input &
So we started by using contrastive learning for the initial representation,
but then we tried to, uh, sort of... actually we shifted to a curriculum
learning approach to stabilize training. We're seeing perplexity drop to about
four point two, which is great, but we're still trying to, how should I put it,
optimize the knowledge distillation process from PaLM two.
We used a teacher model with five hundred forty four billion parameters and...
well, we're basically compressing it into a smaller student, you know, right?
\\

AgenticASR Output &
So we started by using contrastive learning for the initial representation,
but then we shifted to a curriculum learning approach to stabilize training.
We're seeing perplexity drop to about 4.2, which is great, but we're still
trying to optimize the knowledge distillation process from Palm 2.
We used a teacher model with 544 billion parameters, and we're compressing it
into a smaller student, right?
\\

% ================= Customer Service =================

\midrule
\multicolumn{2}{l}{\textbf{Metric: Customer Service}}\\
\midrule

Clean Transcript &
I just noticed something on my statement. Look, I am getting really frustrated here. I think I was charged twice for the same subscription. The same exact charge twice! And then, get this, my credit card was declined but I have sufficient funds. I know I do. I have \$500 in there. I just used it for DoorDash 10 minutes ago and it worked fine. My email is Marcus@email.com. Can you just check why it says I owe you \$249.99? I already paid that. I'm just so tired of this. I keep getting the same answer. I need this fixed now. Please just check the record. My email again is Marcus@email.com. Let me know if you can see the double charge on the system.
\\

Oral Input &
Actually... oh, wait. I just noticed something on my statement. Look, I am getting really frustrated here. I think I was charged twice for the same subscription. Like, the same exact charge twice! And then, get this, my credit card was declined but I have sufficient funds. I know I do. I have like five hundred dollars in there. I mean, I just used it for DoorDash like ten minutes ago and it worked fine. It worked fine! My email is Marcus at email dot com. Can you just check why it says I owe you two hundred forty nine ninety nine? I already paid that. I'm just... I'm just so tired of this. I keep getting the same answer. I need this fixed now. I mean, I really need this fixed now. Please just check the record. My email again is Marcus at email dot com. Let me know if you can see the double charge on the system.
\\

AgenticASR Output &
I just noticed something on my statement. Look, I am getting really frustrated here. I think I was charged twice for the same subscription. The same exact charge twice. And then get this, my credit card was declined, but I have sufficient funds. I know I do. I have \$500 in there. I just used it for DoorDash 10 minutes ago, and it worked fine. My email is markus@email.com. Can you just check why it says I owe you \$249.99? I already paid that. I'm just so tired of this. Keep getting the same answer. I need this fixed now. Please just check the record. My email again is markus@email.com. Let me know if you can see the double charge in the system.
\\

% ================= Daily Chat =================

\midrule
\multicolumn{2}{l}{\textbf{Metric: Daily Chat}}\\
\midrule

Clean Transcript &
People think Camden has bad clinics but actually they're great. Anyway, you oughta see Hana. She used to work in Cape Town and she's the best dentist around here
\\

Oral Input &
um people think Camden has bad clinics but actually they're great... uh we were discussing a favorite childhood cartoon like... wait anyway you oughta see Hana. she used to work in Cape Town and she's like the best dentist around here
\\

AgenticASR Output &
People think Camden has bad clinics, but actually, they're great. Anyway, you ought to see Hannah. She used to work in Cape Town, and she's the best dentist around here.
\\

\bottomrule

\end{tabularx}

\end{table*}

\begin{table*}[t]
\centering

\label{tab:aasr_examples}

\renewcommand{\arraystretch}{1.25}

\small

\begin{tabularx}{0.96\linewidth}{
>{\bfseries}p{2.5cm}
X
}

\toprule

% ================= Dictation =================

\midrule
\multicolumn{2}{l}{\textbf{Metric: Dictation Memo}}\\
\midrule

Clean Transcript &
The guy at 666 Mountain Pass, Salt Lake City, Utah 84101. Need to email client for deposit. I am contacting you about the voice note for lecture, due by end of day tomorrow.
\\

Oral Input &
Uh, for the... wait, the guy at six hundred sixty six Mountain Pass Salt Lake City Utah eight four one zero one. Need... need to email client for deposit. I... I am contacting you about... uh... voice note for lecture. Due by end of day tomorrow.
\\

AgenticASR Output &
For the guy at 666 Mountain Pass, Salt Lake City, Utah 84101, need to email client for deposit. I am contacting you about a voice note for lecture. Due by end of day tomorrow.
\\

% ================= Meeting =================
%230 232
\midrule
\multicolumn{2}{l}{\textbf{Metric: Meeting}}\\
\midrule

Clean Transcript &
I just remembered we need to look at the latest feedback from Ji-won Park in the Marketing team. Anyway, she mentioned that our CAC is way higher than expected. And it looks like our burn rate is exceeding the monthly budget by maybe 12\%? So we really need to focus on cloud spend optimization for AWS. But the main issue is these cross-functional dependency bottlenecks, right? It is just making everything move so slow.
\\

Oral Input &
Actually... oh, I just remembered, uh, we need to look at the latest feedback from... uh... Ji won Park in the Sales team... oh wait, not Sales, Marketing team. Anyway, she mentioned that our CAC is like, way higher than expected. And uh, it looks like our burn rate exceeding monthly budget by, uh, maybe twelve percent? So we really need to focus on cloud spend optimization for AWS. But like, the main issue is these... uh... cross functional dependency bottlenecks, right? It is just making everything move so slow.
\\

AgenticASR Output &
I just remembered. We need to look at the latest feedback from G1 Park in the marketing team. Anyway, she mentioned that our CAC is way higher than expected, and it looks like our burn rate is exceeding the monthly budget by maybe 12\%. So we really need to focus on cloud spend optimization for AWS. But the main issue is these cross-functional dependency bottlenecks, right? It is just making everything move so slow.
\\

% ================= Tech =================

\midrule
\multicolumn{2}{l}{\textbf{Metric: Tech}}\\
\midrule

Clean Transcript &
Hey! So I got this Express app and I need to add some authentication middleware. Currently I'm using Azure SQL Database for user storage. Anyway, I just ran docker-compose up to test locally and it's fine, but I'm seeing some weird cluster-dns issues in the dev environment. I tried a kubectl rollout undo but that didn't help, so I'm gonna try an ArgoCD app sync. Can you help me write the middleware logic?
\\

Oral Input &
Hey! So I got this Express app and I need to add some authentication middleware. Currently I'm using Azure Cosmos DB... oh no, I mean Azure SQL Database for user storage. Anyway, I just ran docker-compose up to test locally and it's fine, but I'm seeing some weird cluster-dns issues in the dev environment. I tried a kubectl rollout undo but that didn't help, so I'm gonna try an argocd app sync. Can you help me write the middleware logic?
\\

AgenticASR Output &
Hey, so I got this Express app and I need to add some authentication middleware. Currently, I'm using Azure SQL database for user storage. Anyway, I just ran Docker Compose up to test locally and it's fine. But I'm seeing some weird cluster DNS issues in the dev environment. I tried a Kubectl rollout undo, but that didn't help. So I'm gonna try an Argo CD app sync. Can you help me write the middleware logic?
\\

% ================= Voice Search =================

\midrule
\multicolumn{2}{l}{\textbf{Metric: Voice Search}}\\
\midrule

Clean Transcript &
Hey, check flight status for flight 120. I thought it was delayed 3.5 hours,
no, it's just a short delay. Check the gate. I was browsing Pinterest and
Duolingo and saw some ad for \$0.88, but that's not it. The flight is
departing from terminal 3. Is it on time for the gate change? I need to know
if it's still at the same gate.
\\

Oral Input &
Hey, check flight status for flight one hundred twenty. I thought it was
delayed three and a half hours, oh no, it's just a short delay. Check check
the gate. I was looking at... I mean, I was browsing Pinterest and Duolingo
and saw some ad for eighty-eight cents, but that's not it. The flight... the
flight is departing from terminal two, no, terminal three. Wait, it's
departing from terminal three.
\\

AgenticASR Output &
Hey, check flight status for flight 120. I thought it was delayed 3.5 hours.
Oh no, it's just a short delay. Check the gate. I was browsing Pinterest and
Duolingo and saw some ad for 88 cents, but that's not it. The flight is
departing from Terminal 3. Is it on time for the gate change? I needed to know
if it's still at the same gate.
\\

\bottomrule

\end{tabularx}

\caption{Examples from different scenarios in AASR-Bench.
Each example contains the clean transcript, oral speech input,
and the refined output generated by AgenticASR.}

\end{table*}

\end{CJK}

\bibliography{bibo}


\end{document}